\documentclass[sigconf]{aamas} 

\usepackage{graphicx}
\usepackage{amsmath}
\usepackage{amsfonts}
\usepackage{commath}
\usepackage{algpseudocode}
\usepackage{algorithm}
\usepackage{amsthm}
\usepackage{caption}
\usepackage{subcaption}
\algnewcommand{\require}[1]{%
  \State \textbf{Require: }\parbox[t]{.8\linewidth}{\raggedright #1}
}
\algnewcommand{\initialize}[1]{%
  \State \textbf{Initialize: }\parbox[t]{.8\linewidth}{\raggedright #1}
}

\DeclareMathOperator*{\argmin}{\arg\!\min}

\newtheorem{theorem}{Theorem}[section]

\newtheorem{lemma}[theorem]{Lemma}

\theoremstyle{remark}

\theoremstyle{definition}
\newtheorem{definition}{Definition}

\usepackage{balance}

\title[AAMAS-2023 Formatting Instructions]{TIDo: Source-free Task Incremental Learning in Non-stationary Environments}

\author{Abhinit Kumar Ambastha}
\affiliation{
  \institution{National University of Singapore}
  \city{Singapore}
  \country{Singapore}}
\email{abhinit@comp.nus.edu.sg}

\author{Leong Tze Yun}
\affiliation{
  \institution{National University of Singapore}
  \city{Singapore}
  \country{Singapore}}
\email{leongty@comp.nus.edu.sg}

\begin{abstract}
This work presents an incremental learning approach for autonomous agents to learn new tasks in a non-stationary environment. Updating a DNN model-based agent to learn new target tasks requires us to store past training data and needs a large labeled target task dataset. Few-shot task incremental learning methods overcome the limitation of labeled target datasets by adapting trained models to learn private target classes using a few labeled representatives and a large unlabeled target dataset. However, the methods assume that the source and target tasks are stationary. We propose a one-shot task incremental learning approach that can adapt to non-stationary source and target tasks. Our approach minimizes adversarial discrepancy between the model's feature space and incoming incremental data to learn an updated hypothesis. We also use distillation loss to reduce catastrophic forgetting of previously learned tasks. Finally, we use Gaussian prototypes to generate exemplar instances eliminating the need to store past training data. Unlike current work in task incremental learning, our model can learn both source and target task updates incrementally. We evaluate our method on various problem settings for incremental object detection and disease prediction model update. We evaluate our approach by measuring the performance of shared class and target private class prediction. Our results show that our approach achieved improved performance compared to existing state-of-the-art task incremental learning methods.
\end{abstract}



\usepackage{colortbl}
\usepackage{graphicx}
\usepackage{tabularx}
\usepackage{footmisc}
\newcommand{\BibTeX}{\rm B\kern-.05em{\sc i\kern-.025em b}\kern-.08em\TeX}

\usepackage{color,soul}

\definecolor{pastel}{HTML}{FAF1E6}
\definecolor{highlightblue}{HTML}{A2DBFA}
\definecolor{pista}{HTML}{CDF0EA}
\definecolor{pinka}{HTML}{FAF3F3}
\definecolor{prettygreen}{HTML}{E4E978}

\begin{document}


\pagestyle{fancy}
\fancyhead{}
\maketitle 

\section{INTRODUCTION}

Task incremental learning problem applies to non-stationary problem settings where an agent needs to update an existing task model but does not have access to large amounts of labeled data. An example of such a setting is an autonomous agent learning an incremental object detection model. Object detection and computer vision models are helpful in various domains, such as robotics, healthcare, e-commerce, and security. A fixed label set and a stationary input data distribution limits a classification model's generalization ability in non-stationary or open-set problem settings. We can overcome these bottlenecks using unsupervised task incremental learning and update the model without access to a large replay memory.

In a task incremental learning problem, the agent aims to learn an optimal hypothesis for both source and target domains \citep{kundu2020class, masana2020class, mi2020generalized, shermin2020adversarial, saito2018maximum}. The source and target are assumed to have undergone a dataset shift \citep{blitzer2006domain}. Hence, their shared class instances have a covariate discrepancy. We assume access to a few labeled target private class instances. The current works address task incremental settings at a single time point. We extend this approach to work in non-stationary settings, where the source and target data are assumed to be available as a dynamic stream. To this effect we propose a new method for task incremental learning -- \textbf{T}ask \textbf{I}ncremental \textbf{Do}main Adaptation (\textbf{TIDo}).
\subsection{BACKGROUND}

In this section, we explore the background topics for this work and theoretical guarantees to learn a task incremental hypothesis using unlabelled data. We first provide a formal definition for task and task incremental learning.

\theoremstyle{definition}
\begin{definition}{\textbf{Task}} A task is defined as a two-tuple $\mathcal{T} = <\mathcal{D},f^{'}>$, where $\mathcal{D}$ is a domain and $f^{'}$ is an approximation of a labeling function for the domain. Learning a task is referred to as learning a close approximation of the aforementioned labeling function.
\label{task_definition}
\end{definition}

\theoremstyle{definition}
\begin{definition}{\textbf{One-shot task incremental learning}}
Given unlabelled target data $x_t^{(t)}\in\mathcal{U}_{t}$ and labelled source domain data $x_s^{(t)}\in\mathcal{D}_{s}$ The target domain label set is given by $C_t$, the shared source and target label set is given by $C_s$, and target-private label set is given by $C^{'}_t$ $C_s = C_t\backslash C^{'}_t$ We have been given a single labelled sample from $C_t^{'}$, $\Tilde{x}_t^{(t)}$. We define task incremental learning as the problem of a target task hypothesis that can predict all labels $C_t$.
\end{definition}

\theoremstyle{definition}
\begin{definition}{\textbf{Hypothesis}} A hypothesis $h \in \mathcal{H}$ refers to an estimate of the labelling function $f:x\to C$, where, $C$ is the label set. The error of a given hypothesis w.r.t. a labelling function for a domain $<\mathcal{D},f>$ is given by: 

\begin{equation}
\epsilon(h,f) := \mathbb{E}_{x\sim\mathcal{D}}[\mathbb{I}(h(x) \neq f(x))],
\end{equation}

where $\mathbb{I}$ is an indicator function.
\label{def:hypothesis}
\end{definition}

For a given source domain,  the true risk of a hypothesis $h\in \mathcal{H}$ is $\epsilon_{S}(h,f)$. Since $\epsilon_{s}(h,f)$ is intractable for most tasks, we use an empirical estimate of the risk, $\hat{\epsilon_{S}}(h,f)$. We assume similar notation for target domain as $\epsilon_{T}(h,f)$ and $\hat{\epsilon_{T}}(h,f)$. The goal is to learn an incremental hypothesis $h^{(t)} \in \mathcal{H}$ at a time point $(t)$, where $\mathcal{H}$ is a hypothesis class.

\begin{equation}
\label{low_risk_obj}
h^{(t)} = \argmin_{h \in \mathcal{H}} \left[ \epsilon^{(t)}_{T}(h,f) + \sum_{i=0}^{t}\epsilon_{s}^{(i)}(h,f) \right]
\end{equation}

We handle the limitation of non-stationary source and target tasks using unsupervised domain adaptation. \cite{blitzer2008learning} show that for a classification task, empirical error and a measure of disagreement between the optimal hypothesis and the proposed hypothesis bounds the true error of a hypothesis. The authors defined the \textit{risk} $\epsilon_s(h,f)$ of a hypothesis $h \in \mathcal{H}$ for a given domain $S$, can be defined as the probability that a hypothesis disagrees with the true labeling function $f$ of a distribution $\mathcal{D}_s$ \citep{blitzer2006domain,blitzer2008learning}:

\begin{equation}
    \epsilon_s(h,f) = \mathbb{E}_{x\sim \mathcal{D}_s}[|h(x) - f(x)|].
\end{equation}

While referring to risk, we use the shorthand $\epsilon_s(h) = \epsilon_s(h,f)$. We use the notation $\hat{\epsilon}_s(h)$ to denote the empirical risk of a hypothesis $h$ for domain $S$.

Blitzer et al. \cite{blitzer2008learning} defined $d_{\mathcal{H}\Delta\mathcal{H}}$ as the measure of maximum disagreement between any hypothesis in a hypothesis class. For a hypothesis space $\mathcal{H}$, $\mathcal{H}\Delta\mathcal{H}$ is defined as a symmetric difference hypothesis space:

\begin{equation}
    \mathcal{H}\Delta\mathcal{H} = {h(x) \oplus h^{'}(x): h,h^{'} \in \mathcal{H}},
\end{equation}

where $\oplus$ is the XOR operator.

$d_{\mathcal{H}\Delta\mathcal{H}}$ was shown to satisfy the following inequality for any hypotheses, $h,h^{'} \in \mathcal{H}$ and domains $S$ and $T$:

\begin{equation}
    |\epsilon_s(h,h^*) - \epsilon_t(h,h^*)| \leq \frac{1}{2}d_{\mathcal{H}\Delta\mathcal{H}}
\end{equation}

\theoremstyle{definition}
\begin{definition}{\textbf{Vapnik-Chervonenkis dimension (VC dimension)}\citep{vapnik1994measuring}}
The Vapnik-Chervonenkis dimension, $VC(\mathcal{H})$, of hypothesis space $\mathcal{H}$ defined over instance space $X$ is the size of the largest finite subset of $X$ shattered by $\mathcal{H}$. If arbitrarily large finite sets of $X$ can be shattered by $\mathcal{H}$ , then $VC(\mathcal{H}) \equiv \infty$
\end{definition}

Lemma \ref{lem:truetargetrisk} shows that we can bind target task risk with source task risk.

\theoremstyle{lemma}
\begin{lemma} 
\label{lem:truetargetrisk}
\citep{blitzer2008learning} For a given source ($S$) and target ($T$) domain, Let $\mathcal{H}$ be a hypothesis class and $h^* \in \mathcal{H}$ be the optimal hypothesis. Let $d_{\mathcal{H}\Delta\mathcal{H}}$ be a symmetric hypothesis space distance. Then for every $h \in \mathcal{H}$ we have,

\begin{equation}
\epsilon_T(h) \leq \epsilon_T(h^*) + \epsilon_T(h,h^*) \leq \epsilon_S(h) + \lambda + \frac{1}{2}d_{\mathcal{H}\Delta\mathcal{H}}(\mathcal{D}_S,\mathcal{D}_T)
\end{equation}

where, 

\begin{equation}
\lambda = \epsilon_{T}(h^*) + \epsilon_{s}(h^*)
\end{equation}

\end{lemma}

In Theorem \ref{thm:target_source}, we show that the risk of an incremental model update using target data will be theoretically bounded by the average risk of the source data provided in the iteration. This ensures a theoretical upper bound of model error when a target dataset is used to update an existing model.

\theoremstyle{theorem}
\begin{theorem} 
\label{thm:target_source}
Let $\mathcal{\hat{D}}_T$ be the empirically estimated target task distribution and $\hat{\mathcal{D}}^{(t)}_{s}$ be the empirical source distribution. Let $d_{\mathcal{H}\Delta\mathcal{H}}$ be a symmetric hypothesis space distance, then for every $h \in \mathcal{H}$, we can show that the true target risk is bound by the average true source risk and the domain discrepancy between $\mathcal{\hat{D}}_T$ and $\hat{\mathcal{D}}^{(t)}_{s}$.

\begin{equation}
\begin{split}
\epsilon_{T}(h) & \leq \frac{1}{t}\sum_{i=1}^{t}
\left(
\epsilon_{S}(h^{(i)}) + \frac{1}{2}d_{\mathcal{H}\Delta\mathcal{H}}(\mathcal{\hat{D}}^{(i)}_{S}, \mathcal{\hat{D}}_{T}) \right) +\lambda^{*(t-1)}
\end{split}
\end{equation}

Where,

\begin{equation}
\label{lambda2}
\lambda^{*(t-1)} =\frac{1}{t}\sum_{i=1}^{t}[\epsilon_{T}(h^{(i-1)}) +  \epsilon_{S}(h^{(i-1)})]
\end{equation}

\end{theorem}

In Theorem \ref{thm:target_target}, We show that the risk of an incremental model update is theoretically bounded by the average risk of the previously introduced target data and domain discrepancy between the source and target domains. This shows that the model should be able to learn incrementally using new domain data as long as it has low domain discrepancy compared to the original model source data.

\theoremstyle{theorem}
\begin{theorem}
\label{thm:target_target}
Let $\mathcal{D}_T$ be the true target task distribution and $\mathcal{D}^{(t)}_{s}$ be the true source distribution. Let $d_{\mathcal{H}\Delta\mathcal{H}}$ be a symmetric hypothesis space distance. For every $h \in \mathcal{H}$, we can show that the true target risk is bounded by the average true target risk of the previous increments, the domain discrepancy between $\mathcal{\hat{D}}_T$ and $\hat{\mathcal{D}}^{(t)}_{s}$ and the optimal hypothesis risk $\lambda^{*(t-1)}$ of the existing model.

\begin{equation}
\epsilon_T(h^{(t)}) \leq \frac{1}{t}\left(\sum_{i=1}^{t}\epsilon_T(h^{(i-1)}) + \frac{1}{2t}\sum_{i=1}^{t}d_{\mathcal{H}\Delta\mathcal{H}}(\mathcal{D}_S,\mathcal{D}_T)\right) + \lambda^{*(i-1)}
\end{equation}

Where,

\begin{equation}
\label{lambda2}
\lambda^{*(t-1)} =\frac{1}{t}\sum_{i=1}^{t}[\epsilon_{T}(h^{(i-1)}) +  \epsilon_{S}(h^{(i-1)})]
\end{equation}

\end{theorem}

In Theorem \ref{thm:target_empsource}, we can learn an incremental model by reducing the empirical $H-$distance (a measure of domain divergence) between unlabelled source and target domain data. The theorem is an incremental extension of the work by Blitzer et al. \cite{blitzer2006domain,blitzer2008learning}.

\theoremstyle{theorem}
\begin{theorem} 
\label{thm:target_empsource}
Let $\mathcal{\hat{U}}^{(t)}_{T}$ be the empirically estimated unlabelled target distribution and $\mathcal{\hat{U}}^{(t)}_{s^{(t)}}$ be the empirical unlabelled source empirical distribution. Let $d_{\mathcal{H}\Delta\mathcal{H}}$ be a symmetric hypothesis space distance. $m^\prime_{i}$ is the size of the unlabelled target and source samples, and $d$ is the Vapnik–Chervonenkis dimension of the current hypothesis. Then for every $h \in \mathcal{H}$ for a probability at least $1-\delta$, 

\begin{multline}
\epsilon_{T}(h^{(t)}) \leq  \frac{1}{t} \sum_{i=1}^{t} \left(
\hat{\epsilon}_{S^{(i)}}(h^{(i)})
+ \frac{1}{2}d_{\mathcal{H}\Delta\mathcal{H}}(\mathcal{\hat{U}}^{(t)}_{S^{(i)}}, \mathcal{\hat{U}}^{(t)}_{T})\right) \\
+ \frac{1}{t} \sum_{i=1}^{t}\left(4\sqrt{\frac{2d\log(2m^\prime_{i}) + \log(\frac{4}{\delta})}{m^\prime_{i}}}\right) + \lambda^{*(t-1)}
\end{multline}

Where,

\begin{equation}
\label{lambda2}
\lambda^{*(t-1)} =\frac{1}{t}\sum_{i=1}^{t}[\epsilon_{T}(h^{(t)}) +  \epsilon_{S}(h^{(t)})]
\end{equation}

\end{theorem}
\section{RELATED WORKS}

This section explores current works used to learn an autonomous task incremental learning agent.

Li et al. \cite{li2017learning} propose a neural network-based approach (Learning without forgetting) to carry out task incremental learning with minimal increase in parametric space size while satisfying low data resource conditions. The proposed method learns new target private class mappings by adding new neurons to the output layer of a classification network. The goal of the approach is to retain the classification performance for the previous tasks while \textit{incrementally} learning new tasks or classes. The authors use distillation loss \cite{hinton2015distilling} to minimize catastrophic forgetting.

Rebuffi et al. \cite{rebuffi2017icarl} propose a supervised incremental learning approach (Incremental Classifier and Representation Learning) that uses nearest mean matching and class-wise representatives from the input data. The authors update the representative instance sets (referred to as \textit{exemplars}) using samples from the incremental input data batches. In our work, we address an unsupervised source-free approach to overcome the limitation of storing representative examples and the need to label incoming incremental data.

Hoffman et al. \cite{hoffman2014continuous} provide a supervised approach (continual manifold adaptation) to learn a low dimensional embedding subspace for incoming target data. The work update parametric kernels to model an evolving target task distribution. Using a kernel-based approach is computationally intensive if the target dataset size is large, which limits the scalability of the approach.

Kundu et al. \citep{kundu2020class} propose a source-free class incremental learning approach that updates a model in a non-stationary environment. The authors provide a way to learn a target model with private and shared classes but assume known target classes at the time of incremental domain adaptation. Our work is an incremental extension of this work. Also, the work addresses domain shift compensation using L2 regularization, which fails to account for unknown classes. We address these challenges using distillation loss to accommodate future target private classes and an adversarial domain confusion loss to minimize domain shift for a non-stationary target domain.

We compared our approach to unsupervised domain adaptation methods. Ganin et al. propose the domain adversarial neural network (DANN) \citep{ganin2016domain, zhao2018adversarial}. Domain adaptation methods cannot compensate for non-stationary source distribution and do not provide the ability to add target private classes (open-set problem setting). We compare our work with DANN combined with a target private classifier. Due to the few labeled samples available for target private classes, it cannot learn an optimal hypothesis and has low predictive accuracy.
\section{OUR APPROACH}

\begin{figure}[ht]
    \centering
    \includegraphics[width=0.47\textwidth]{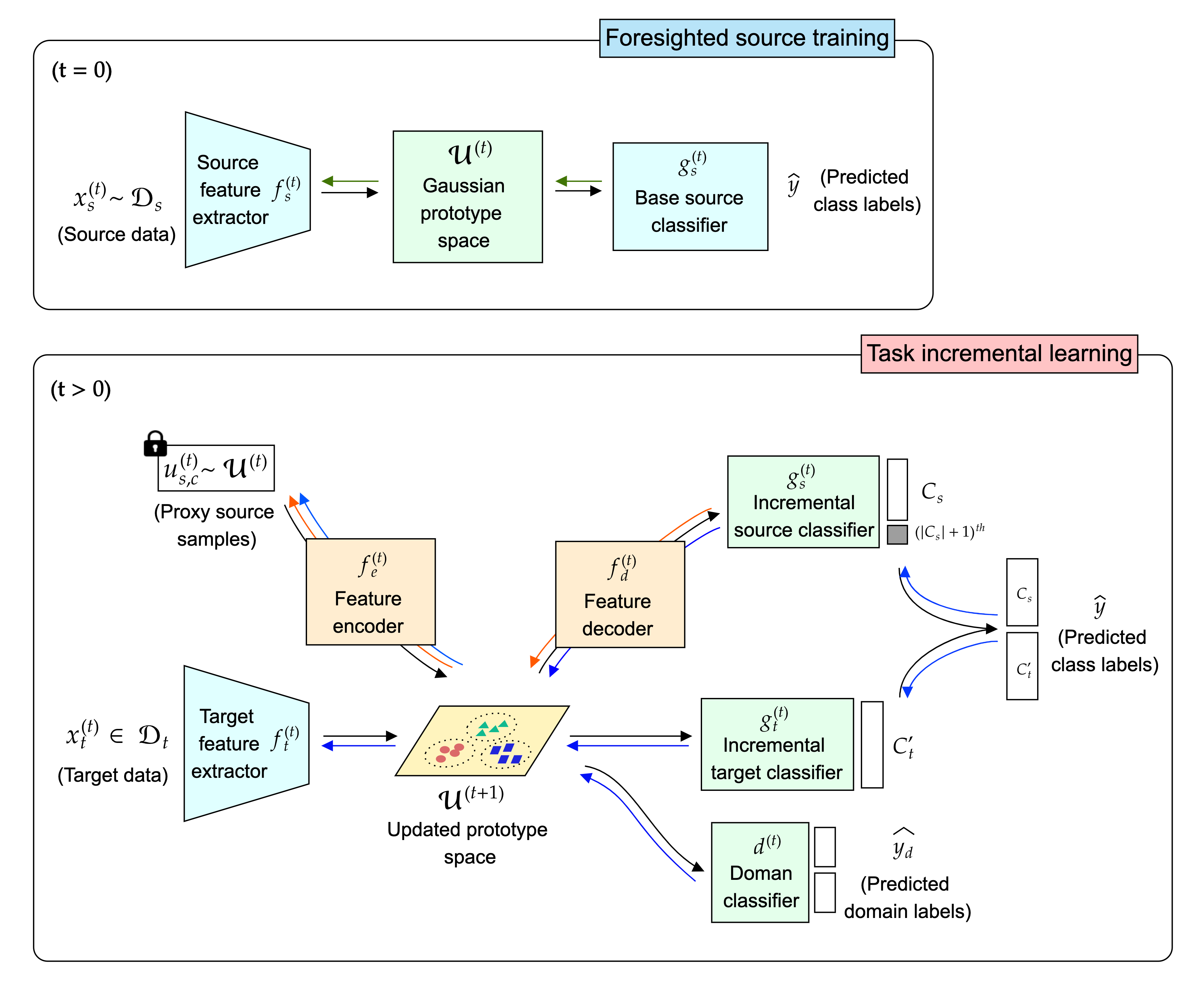}
    \caption{\textbf{TIDo architecture:} Proposed method architecture for task incremental learning architecture.}
    \Description{\textbf{TIDo architecture:} Proposed method architecture for task incremental learning architecture.}
    \label{fig:tido}
\end{figure}

Our approach is divided into two stages -- \textit{foresighted learning} and \textit{task incremental update}. In the foresighted learning stage, an agent learns a generative model of the source data feature space. Foresighted learning helps the agent to generate representative samples of past data for future incremental model updates. In the task incremental learning stage, the agent updates its internal model state using unlabeled target data and a single labeled sample for target private classes.

\subsection{Foresighted learning} 

This section describes the foresighted learning stage. This stage aims to identify tight class-wise clusters in feature posterior distribution using Gaussian estimation.

We denote the feature extractor function as $f_s$ and the classifier function as $g_s$, which maps the feature extractor output to a $|C_{s}+1|$-class label space (where $C_s$ is the source task label set size). The latent space is denoted by $\mathcal{U}$. We minimize cross-entropy loss ($l_{ce}$) to learn $g_s$.
\begin{equation}
    l_{ce} = \mathop{\mathbb{E}}_{(x_s,y_s)\sim \mathcal{D}^{(t)}_{s}} l_{ce}(g_s \cdot f_s (x_s), y_s)
    \label{vanilla_ce}
\end{equation}
Cross-entropy loss ensures discriminative decision boundaries in the latent feature space but leads to over-confident predictions. To generate representative samples for source distribution for future iterations, we minimize category bias by penalizing over-confident prediction. We achieve this by identifying out-of-distribution (OOD) samples. Re-using the trained base model to classify unknown classes leads to \textit{negative learning}, \textit{i.e.}, and misclassification of instances belonging to unknown classes as one of the known classes. This is due to the inherent generalization bias of the source model. Kundu et al. \citep{kundu2020class} suggest detecting OOD instances to identify instances belonging to unknown classes. This is based on the understanding that instances from unknown classes lie in low-density regions of the instances of the shared classes. Kundu et al. \citep{kundu2020class} achieve this by mapping the source instances to a latent space with an underlying global prior distribution given by $\mathcal{N}(\mu,\sigma)$. Next, the instances from the target domain which lie beyond the $3\sigma$ range were considered to belong to unknown classes.

We use a class separability objective $\mathcal{L}_{s1}$ to enforce the class-wise features to attain higher affinity to the class-wise prototypes.

\begin{equation}
    \mathcal{L}_s = \mathcal{L}_{s1} + \mathcal{L}_{s2}
\end{equation}
\begin{equation}
    \mathcal{L}_{s1}: \mathop{\mathbb{E}}_{(x_s,y_s)\sim \mathcal{D}_{s}} - \log \left[ \frac{\exp(\mathcal{P}_s^{y_s}(u_s))}{\sum_{c\in C_s}\exp(\mathcal{P}_s^{c}(u_s))} \right]
\end{equation}
\begin{equation}
\begin{split}
    \mathcal{L}_{s2}: \mathop{\mathbb{E}}_{(x_s,y_s)\sim \mathcal{D}_{s}} l_{ce}(\sigma(g_s^{(t)}\dot f_s^{(t)}(x_s),\tau),y_s) \\
 + \mathop{\mathbb{E}}_{(u_n,y_n)\sim \mathcal{D}_{n}} l_{ce}(\sigma(g_s^{(t)}(u_n),\tau),y_n)
\end{split}
\end{equation}
Where $\sigma$ denotes distillation soft loss \citep{hinton2015distilling},
\begin{equation}
\sigma(\mathbf{z},\tau) = \frac{e^{\frac{z}{\tau}}}{\sum e^{\frac{z}{\tau}}}
\end{equation}

$\mathcal{D}_{n}$ is the distribution of the negative samples, and $(u_n,y_n)$ represents the negative samples with $y_n$ being the $(|C_s|+1)^{th}$ class. Since we don't need distillation loss for this stage, we set $\tau = 1$.

\subsection{Task incremental update}

In this section, we describe the domain incremental update
stage of the proposed method. We use the learned prototypes and the unlabeled target domain data to incrementally update the base classifier for the target task. In this stage, we use the $\mathcal{U}-$space guides as shared class cluster centroids and single target private samples as the target private class cluster centroids. We use an encoder-decoder approach to fine-tune the $\mathcal{U}-$space to accommodate target private guides. This way, we learn a $\mathcal{U}^{th}-$space which is used to represent previous iteration samples.

Several discrepancy metrics have been proposed to match the moments of the shared class instances from source and target distributions. Adversarially trained domain discriminators are used to reducing the empirical hypothesis distance ($d_{\mathcal{H}\nabla\mathcal{H}}$) between the source and target distributions, which has been shown to reduce the distance between the source and target distributions.

We learn the guides for the target domain, $\mathcal{U}^{(t+1)}$ using the source prototype space $\mathcal{U}$. Using fixed source guides for target space reduces flexibility in accommodating target private classes. We initialize the $\mathcal{U}^{(t+1)}$ guides: $v_g^c = f_e(\mu_s^c) \forall c\in C_s$ and $v_g^c = \hat{x}_t^{(t)} \forall c\in C^{'}_t$ and calculate confident samples $\mathcal{B}_{t}^c$ which are pseudo-labelled using the guides ($k$). We use a domain projection auto-encoder to enable \textit{mobility} of guides explicitly. The target domain contains private class instances which position themselves in the low-density regions of the $\mathcal{U}^{t+1}-$space. We use a reconstruction loss and L2-norm to maintain the previously learned source guide space ($\mathcal{U}-$space) semantics. By training the auto-encoder layers using the gradient from the classifier and domain discriminator, we adversarially train $\mathcal{U}^{(t+1)}-$space.

The $\mathcal{U}^{(t+1)}$ guides are aligned using the adversarial domain confusion loss:

\begin{equation}
    \mathcal{L}_{d}: -d_{\mathcal{H}\nabla \mathcal{H}}(v_t,v_g^c)
\end{equation}

In order to learn an efficient domain projection $f_e:\mathcal{U}\rightarrow\mathcal{U}^{(t+1)}$ and $f_d:\mathcal{U}^{(t+1)}\rightarrow\mathcal{U}$ we use reconstruction error similar to an auto-encoder. We also use distillation loss with $\tau = 2$ to ensure low catastrophic forgetting for the previously learned shared classes.
\begin{equation}
    \mathcal{L}_{r} = \mathcal{L}_{r1} + \mathcal{L}_{r2}
\end{equation}
\begin{equation}
    \mathcal{L}_{r1}: \mathop{\mathbb{E}}_{(u_s^c)\sim \mathcal{P}_{s}^c} l_{ce}(\sigma(\hat{y}(u_s^c),\tau),c)
\end{equation}
\begin{equation}
    \mathcal{L}_{r2}: \mathop{\mathbb{E}}_{(u_s^c)\sim \mathcal{P}_{s}^c} l_{2}(f_d\cdot f_e(u_s^c),u_s^c)^2
\end{equation}
To learn new target private classes, we apply cross-entropy loss to target confident samples:
\begin{equation}
    \mathcal{L}_c: \mathop{\mathbb{E}}_{(x_t)\sim \mathcal{B}_{t}^c} l_{ce}(\hat{y}(v_t),c), \forall c\in C_t
\end{equation}
\section{Algorithm}

Algorithm~\ref{alg:TaskIncrementalLearning} outlines the task incremental update implementation. We initialize the source generative distribution using the prototypes from the previous stage (line~\ref{task:init}). To enable the mobility of guides, we train an auto-encoder network $f_d(f_e(\cdot))$ (line~\ref{task:ae_start}-\ref{task:ae_end}). We use an $L2$-norm loss as a reconstruction error to train the auto-encoder.

Since we want the incremental learning agent to learn new classes from the target task, we need to update the guides to include new target class cluster guides. To this effect, we use a single labeled instance (assumed to be available) from each target task class as target class guides (line~\ref{task:guide_init}). 

\begin{algorithm}[!ht]
\caption{Task Incremental Learning algorithm}
\begin{algorithmic}[1]
\require{Target samples $\mathcal{D}_t$, Gaussian Prototypes $\mathcal{P}_s^c$, model parameters $\theta_{f_s^{(t)}}, \theta_{g_s^{(t)}}, \theta_{f_t^{(t)}}, \theta_{g_t^{(t)}}, \theta_{f_e^{(t)}}, \theta_{f_d^{(t)}}, \theta_{d^{(t)}}$, training sample size $N$, percentage of confident samples $n$}
\initialize{$\theta_{f_t^{(t)}} \gets \theta_{f_s^{(t)}}$} \label{task:init}
\Repeat \label{task:ae_start}
    \State Obtain a mini-batch of proxy-source samples $S=\{\mathbf{u}_s^c\sim \mathcal{P}_s^c:c\in \mathcal{C}_s\}$
    \State $\theta_{f_e^{(t)}} \gets \theta_{f_e^{(t)}} + \text{Adam}_{\{f_e^{(t)}\}}(-\nabla\frac{1}{|S|}\sum_{\mathbf{u}_s^c\in S}l_2(u_s^c,f_e(u_s^c))^2)$
    \State $\theta_{f_d^{(t)}} \gets \theta_{f_d^{(t)}} + \text{Adam}_{\{f_d^{(t)}\}}(-\nabla\frac{1}{|S|}\sum_{\mathbf{u}_s^c\in S}l_2(u_s^c,f_d(u_s^c))^2)$
\Until{Convergence} \label{task:ae_end}
\State \textbf{Loss} $\gets [\mathcal{L}_{r1}, \mathcal{L}_{r2},\mathcal{L}_c,\mathcal{L}_{d}]$
\State \textbf{Opt} $\gets [\text{Adam}_{\{f_e^{(t)},f_d^{(t)},g_t^{(t)}\}}, \text{Adam}_{\{f_e^{(t)},f_d^{(t)}\}}, \text{Adam}_{\{f_t^{(t)},g_t^{(t)}\}},$ \\
\hspace{3.5em} $\text{Adam}_{\{f_t^{(t)}\}}, \text{Adam}_{\{f_e^{(t)},f_t^{(t)}\}}]$
\Repeat
    \State \textit{iter} $\gets$ \textit{iter}$+1$, \textit{cur} $\gets$ \textit{iter mod }$5$
    \State $\mathbf{v}_g^c \gets f_e(\mathbf{\mu}_s^c) \forall c\in \mathcal{C}_s, \mathbf{v}_g^c \gets f_t(\Tilde{x}_t^c) \forall c\in \mathcal{C}^{'}_{t}$ \label{task:guide_init}
    \For{$u_s^c \sim \mathcal{P}_s^c: c\in \mathcal{C}_s$} \label{task:class_start}
        \State $v_s^c \gets f_e(\mathbf{u}_s^c); \hat{u}_s^c \gets f_d(v_s^c)$; 
        \State $\hat{y} \gets g_s(\hat{u}_s^c)|_{c \in \mathcal{C}_s} \Vert g_t(v_s^c)$
        \State $\mathcal{L}_{r1} + l_{mse}(\hat{u_s^c},u_s^c)$
        \State $\mathcal{L}_c \gets \mathcal{L}_c + l_{ce}(\sigma(\hat{y}_s),c)$
    \EndFor \label{task:class_end}
    \For{$\mathbf{x}_t \in \{\mathbf{x}_t \sim \mathcal{D}_t\}$} \label{task:target_prediction_start}
        \State $v_t \gets f_t(\mathbf{x}_t); \mathbf{u}_t\gets f_d(v_t); \hat{y}_t\gets g_s(\hat{u}_t)|_{c\in \mathcal{C}_s} \Vert g_t(v_t)$ 
        \State $d\gets \text{min}_{c\in \mathcal{C}_t}l_2(v_t,v_g^c); k\gets \text{arg min} (d)$
        \State $\mathcal{L}_{r2} \gets \mathcal{L}_{r2} + l_{mse}(\mathbf{u}_t,\hat{v_t})$ \label{task:target_recon_loss}
    \EndFor \label{task:target_prediction_end}
    \For{$u_s^c \sim \mathcal{P}_s^c$, $\mathbf{x}_t \in \{\mathbf{x}_t \sim \mathcal{D}_t\}$} \label{task:d_start}
        \State $v \gets f_t(\mathbf{x}_t)$; $\hat{y_d} \gets d([u_s^c,v])$;
        \State $\mathcal{L}_d \gets \mathcal{L}_d + l_{ce}(\hat{y_d},[0,1])$
    \EndFor \label{task:d_end}
    \If{\textit{reached the end of an epoch}} \label{task:update_start}
        \State UpdateTaskIncrementalGradients(\textbf{Loss},\textbf{Opt}) \label{task:update}
        \State Label samples in $\mathcal{D}_t$ using \textit{guides} $\{v_g^c:c\in \mathcal{C}_t\}$
        \State $\mathcal{P}_t^c \gets$ Gaussian Prototypes obtained using pseudo-label target samples \label{task:guide_update}
    \EndIf \label{task:update_end}
\Until{Convergence}
\end{algorithmic}
\label{alg:TaskIncrementalLearning}
\end{algorithm}

In line~\ref{task:class_start}-\ref{task:class_end}, we fine-tune the feature extractor network using samples from the class-wise prototype distributions. We assume an open-set problem setting, and the target domain data is assumed to contain instances from the source domain. To learn a single classifier for source and target tasks, we pass the unlabeled target instances and source domain samples to both the source domain classifier and target domain classifier. In line~\ref{task:target_prediction_start}-\ref{task:target_prediction_end}, we obtain the pseudo-labels for the target domain data along with the predictions from the updated joint classifier.

To align the source and target domain distributions, we use a domain discriminator network, which trains the feature extractor adversarially along with the joint classifier loss (line~\ref{task:d_start}-\ref{task:d_end}). Finally, we update the parameters of all the components of the task incremental network and update the prototypes (line~\ref{task:update_start}-\ref{task:update_end}). The updated prototypes will serve as the source prototypes in the next iteration, along with the new source domain (if any) to generate the samples. The gradient update algorithm (algorithm~\ref{alg:UpdateTaskIncrementalGradients}) provides the gradient update step for the task incremental learning network components.

\begin{algorithm}[ht]
\caption{Gradient update algorithm}
\begin{algorithmic}[1]
\require{\textbf{Model parameters}, \textbf{Loss}, \textbf{Opt}}
\State $\theta_{f_t^{(t)}} \gets \theta_{f_t^{(t)}} + \text{Adam}_{\{f_t^{(t)}\}}(-\nabla\frac{1}{N}\sum \mathcal{L}_c)$
\State $\theta_{d^{(t)}} \gets \theta_{d^{(t)}} - \text{Adam}_{\{d^{(t)}, f_t^{(t)}\}}(-\nabla\frac{1}{N}\sum \mathcal{L}_d)$
\State $\theta_{f_t^{(t)}} \gets \theta_{f_t^{(t)}} - \text{Adam}_{\{d^{(t)}, f_t^{(t)}\}}(-\nabla\frac{1}{N}\sum \mathcal{L}_d)$
\State $\theta_{f_t^{(t)}} \gets \theta_{f_t^{(t)}} - \text{Adam}_{\{ f_t^{(t)}\}}(-\nabla\frac{1}{N}\sum \mathcal{L}_{r1})$
\State $\theta_{f_e^{(t)}} \gets \theta_{f_e^{(t)}} + \text{Adam}_{\{f_e^{(t)},f_d^{(t)}\}}(-\nabla\frac{1}{N}\sum \mathcal{L}_{r1})$
\State $\theta_{f_d^{(t)}} \gets \theta_{f_d^{(t)}} + \text{Adam}_{\{f_e^{(t)},f_d^{(t)}\}}(-\nabla\frac{1}{N}\sum \mathcal{L}_{r1})$
\State $\theta_{f_t^{(t)}} \gets \theta_{f_t^{(t)}} + \text{Adam}_{\{ f_t^{(t)}\}}(-\nabla\frac{1}{N}\sum \mathcal{L}_{r2})$
\State $\theta_{f_d^{(t)}} \gets \theta_{f_d^{(t)}} + \text{Adam}_{\{f_d^{(t)}\}}(-\nabla\frac{1}{N}\sum \mathcal{L}_{r2})$
\end{algorithmic}
\label{alg:UpdateTaskIncrementalGradients}
\end{algorithm}

\section{EXPERIMENTS AND RESULTS}

\subsection{Incremental object detection}
We evaluated our proposed method to develop an agent to learn an incremental object detection task. Object detection in real-world images has been used as a benchmark task for several computer vision problems. In order to evaluate our approach, we created an incremental learning task that requires learning a target domain classification model given an initial source domain dataset. 

We use an imaging dataset with multiple object classes and multiple domains. We select one of the domains as the initial labeled source domain while the rest are considered unlabelled target domains. Our goal is to learn a common model for all the domains observed by the model.

\subsubsection{Dataset}
We used the office-31 object recognition dataset \citep{saenko2010adapting} which contains 4652 images from 3 domains and 31 classes. The domains of the dataset are web (Amazon), DSLR, and webcam. The domain details are as follows:

\begin{itemize}
    \item Amazon (A): These are images taken from Amazon \citep{amazon.com_2022}. They are mostly taken in a studio setting with a clear background and standardized lighting. We have an average of 90 images per class.
    \item Digital single-lens reflex camera (D): This domain contains high-resolution images with a pixel resolution of ($4288\times2848$). Each class contains images of 5 objects taken from 3 different angles each. In total, the domain dataset has 423 images.
    \item Webcam (W): This domain contains low-resolution poor-lighting images with a pixel resolution of ($640\times480$). The dataset contains 5 objects per class with 3 angle images each. In total, we have 795 images. These images show considerable noise and color as well as white balance artifacts.
\end{itemize}

The 31 categories are desk lamp, computer, tile cabinet, backpack, bike, bike helmet, mouse, mug, notebook, pen, phone, printer, bookcase, bottle, calculator, desk chair, headphones, keyboard, laptop, letter tray, mobile phone, monitor,  projector, puncher, ring binder, ruler, scissors, speaker, stapler, tape, and trash can.

\begin{table}[ht]
\centering
\caption{Incremental object detection learning task for evaluating task incremental learning methods.}
\label{office31_incremental}
\begin{tabular}{@{}ll@{}}
\toprule
\textbf{Index} &
  \textbf{Inputs} \\ \midrule
$t+0$ &
  \begin{tabular}[c]{@{}l@{}}\textit{Source}: desk lamp, computer, cabinet, backpack, bike\\ \textit{Target}: desk lamp, computer, cabinet, backpack, bike, \\ bike helmet, mouse, mug, notebook, pen\end{tabular} \\
$t+1$ &
  \begin{tabular}[c]{@{}l@{}}\textit{Source}: phone, printer, bookcase\\ \textit{Target}: phone, printer, bookcase, bottle, calculator\end{tabular} \\
$t+2$ &
  \begin{tabular}[c]{@{}l@{}}\textit{Source}: desk chair, headphones, keyboard, laptop, tray\\ \textit{Target}: desk chair, headphones, keyboard, laptop, tray, \\ mobile phone, monitor, projector\end{tabular} \\
$t+3$ &
  \begin{tabular}[c]{@{}l@{}}\textit{Source}: ruler, scissors, speaker, stapler \\ \textit{Target}: ruler, scissors, speaker, stapler, tape, trash can\end{tabular} \\
$t+4$ &
  \textit{Source}: $\varnothing$, \textit{Target}: puncher, ring binder \\ \bottomrule
\end{tabular}%
\end{table}

To evaluate the response of our approach to both open-set differences between the source and target domains and non-stationary source and target domains, we structure the experiment as follows

\begin{itemize}
    \item The domains are introduced incrementally to the model, and the data from the domains (belonging to the same class) is assumed to be sampled from a single non-stationary distribution
    \item In every iteration we introduce a set of shared classes $C_s^{t}$ and target private classes $C_{t}^{'(t)}$
    \item The foresighted learning network learns the source guides every time a new labeled source domain is introduced. For the $l^{th}$ iteration, $f_s^{(t+l)}$ is trained using data sampled from combined data from $u_s^{(t)}$ and $x_s^{(t+l)}$
    \item The target data in every iteration assumed to contain at least one target private class (\textit{i.e.} $C^{'(t+1)} \neq \varnothing$)
\end{itemize}

Like our method, iCARL and CIDA use prototype learning to enable source-free incremental learning. Although this is one of the desiderata of incremental learning, iCARL requires labeled target data. This makes it unsuitable for direct application to the unsupervised task incremental learning problem setting. Our work is motivated by CIDA, and we aim to improve upon the existing method by using an adversarial domain discrepancy estimation instead of the previously proposed alignment loss \citep{kundu2020class}. We also extend it to an incremental learning context. DANN and CMA provide a way to carry out unsupervised learning. We compare our approach to the aforementioned methods to evaluate the efficiency of end-to-end trainable adversarial methods for task incremental learning. 

We evaluate our approach using the incremental learning task outlined in table \ref{office31_incremental}. We evaluate the performance of a given approach at every time point using total accuracy and target private class accuracy. We compare our proposed approach (TIDo) to unsupervised domain adaptation methods (DANN \citep{tzeng2017adversarial}), class incremental domain adaptation methods (iCARL \citep{rebuffi2017icarl}, CIDA \citep{kundu2020class}) and continual learning methods (LwF-MC \citep{li2017learning}, CMA \citep{hoffman2014continuous}). For methods without a provision to incrementally add new classes, we trained a target private classifier (TPC). CIDA-C refers to storing and using combined target task data from past increments; this makes this a pseudo-incremental learning approach.

For $(t+4)^{th}$ iteration of the experiment (refer to table.\ref{office31_incremental}), we have no source dataset. We do not update the source classifier for the DANN approach for this iteration as the approach requires source data to update. Also, iCARL and LwF-MC methods are supervised methods and require labeled target data. We use 5\% labeled samples (available to the rest of the methods for few-shot learning) to serve as the labeled target data.

\begin{table}[ht]
\centering
\caption{Incremental disease prediction learning task for evaluating task incremental learning.}
\label{office31_incremental}
\begin{tabular}{@{}ll@{}}
\toprule
\textbf{Index} & \textbf{Inputs}                                                      \\ \midrule
$t+0$          & \textit{Source}: CN, AD,  \textit{Target}: CN, MCI, AD        \\
$t+1$          & \textit{Source}: CN, MCI, AD,   \textit{Target}: CN, MCI, AD   \\
$t+2$          & \textit{Source}: $\varnothing$,   \textit{Target}: EMCI        \\
$t+3$          & \textit{Source}: $\varnothing$, \textit{Target}: AD, CN, MCI \\ \bottomrule
\end{tabular}%
\end{table}

\begin{table*}[ht]
\centering
\caption{Office-31 incremental object recognition task: comparison of our proposed method (TIDo) with existing incremental learning, continual learning, and unsupervised domain adaptation methods for a few-shot labeled target (5\%), unlabelled target domain data, and labeled source domain data.}
\label{office31_tido}
\begin{tabular}{@{}ccccccccccccc@{}}
\toprule
 &
  \multicolumn{12}{c}{A$\rightarrow$W$\rightarrow$D} \\ \midrule
 &
  \multicolumn{2}{c}{DANN-TPC} &
  \multicolumn{2}{c}{iCARL} &
  \multicolumn{2}{c}{CMA-TPC} &
  \multicolumn{2}{c}{CIDA-C} &
  \multicolumn{2}{c}{LwF-MC} &
  \multicolumn{2}{c}{TIDo} \\ \midrule
Index &
  All (\%) &
  Priv (\%) &
  All (\%) &
  Priv (\%) &
  All (\%) &
  Priv (\%) &
  All (\%) &
  Priv (\%) &
  All (\%) &
  Priv (\%) &
  All (\%) &
  \textit{Priv (\%)} \\ \midrule
$t+0$ &
  57.17 &
  15.91 &
  72.91 &
  48.21 &
  61.22 &
  28.01 &
  \textbf{77.12} &
  72.92 &
  62.12 &
  39.11 &
  75.82 &
  \textbf{75.12} \\
$t+1$ &
  54.12 &
  12.67 &
  75.01 &
  51.23 &
  65.18 &
  27.43 &
  72.23 &
  70.27 &
  62.25 &
  35.24 &
  \textbf{72.81} &
  \textbf{73.63} \\
$t+2$ &
  45.12 &
  19.01 &
  64.21 &
  43.91 &
  56.92 &
  34.22 &
  70.14 &
  69.22 &
  54.50 &
  31.12 &
  \textbf{71.12} &
  \textbf{72.22} \\
$t+3$ &
  40.13 &
  20.87 &
  63.34 &
  43.31 &
  50.85 &
  26.75 &
  69.92 &
  69.91 &
  54.23 &
  30.03 &
  \textbf{70.75} &
  \textbf{71.29} \\
$t+4$ &
  38.23 &
  31.01 &
  60.23 &
  43.33 &
  44.23 &
  29.65 &
  \textbf{73.29} &
  \textbf{78.01} &
  52.15 &
  35.01 &
  72.23 &
  72.16 \\ \midrule
 &
  \multicolumn{12}{c}{D$\rightarrow$A$\rightarrow$W} \\ \midrule
 &
  \multicolumn{2}{c}{DANN-TPC} &
  \multicolumn{2}{c}{iCARL} &
  \multicolumn{2}{c}{CMA-TPC} &
  \multicolumn{2}{c}{CIDA-C} &
  \multicolumn{2}{c}{LwF-MC} &
  \multicolumn{2}{c}{TIDo} \\ \midrule
Index &
  All (\%) &
  Priv (\%) &
  All (\%) &
  Priv (\%) &
  All (\%) &
  Priv (\%) &
  All (\%) &
  Priv (\%) &
  All (\%) &
  Priv (\%) &
  All (\%) &
  \textit{Priv (\%)} \\ \midrule
$t+0$ &
  51.66 &
  20.02 &
  73.22 &
  53.40 &
  68.56 &
  37.12 &
  85.63 &
  82.91 &
  75.34 &
  55.27 &
  \textbf{88.26} &
  \textbf{84.92} \\
$t+1$ &
  50.23 &
  19.10 &
  72.81 &
  51.23 &
  66.30 &
  34.16 &
  81.64 &
  79.22 &
  76.26 &
  56.72 &
  \textbf{84.54} &
  \textbf{82.76} \\
$t+2$ &
  44.66 &
  21.91 &
  71.81 &
  52.42 &
  52.86 &
  21.96 &
  \textbf{76.72} &
  72.01 &
  64.48 &
  53.29 &
  74.86 &
  \textbf{73.66} \\
$t+3$ &
  43.36 &
  17.64 &
  71.74 &
  56.81 &
  51.47 &
  18.58 &
  \textbf{72.12} &
  \textbf{70.26} &
  64.43 &
  44.02 &
  70.49 &
  69.06 \\
$t+4$ &
  41.26 &
  24.63 &
  75.41 &
  65.19 &
  54.63 &
  28.43 &
  \textbf{76.72} &
  \textbf{73.03} &
  65.6 &
  47.7 &
  71.53 &
  70.44 \\ \midrule
 &
  \multicolumn{12}{c}{W$\rightarrow$D$\rightarrow$A} \\ \midrule
 &
  \multicolumn{2}{c}{DANN-TPC} &
  \multicolumn{2}{c}{iCARL} &
  \multicolumn{2}{c}{CMA-TPC} &
  \multicolumn{2}{c}{CIDA-C} &
  \multicolumn{2}{c}{LwF-MC} &
  \multicolumn{2}{c}{TIDo} \\ \midrule
Index &
  All (\%) &
  Priv (\%) &
  All (\%) &
  Priv (\%) &
  All (\%) &
  Priv (\%) &
  All (\%) &
  Priv (\%) &
  All (\%) &
  Priv (\%) &
  All (\%) &
  \textit{Priv (\%)} \\ \midrule
$t+0$ &
  57.29 &
  20.31 &
  75.4 &
  65.2 &
  67.22 &
  18.20 &
  \textbf{84.82} &
  82.22 &
  74.99 &
  67.48 &
  83.92 &
  \textbf{83.20} \\
$t+1$ &
  58.25 &
  18.17 &
  74.59 &
  64.29 &
  67.18 &
  19.02 &
  79.17 &
  82.61 &
  73.62 &
  68.01 &
  \textbf{81.18} &
  \textbf{83.19} \\
$t+2$ &
  56.72 &
  37.45 &
  76.25 &
  77.50 &
  69.93 &
  28.21 &
  \textbf{85.14} &
  \textbf{82.22} &
  67.91 &
  56.02 &
  78.03 &
  80.21 \\
$t+3$ &
  54.29 &
  26.22 &
  76.43 &
  77.02 &
  65.18 &
  24.59 &
  \textbf{82.91} &
  \textbf{79.81} &
  65.23 &
  51.43 &
  77.78 &
  77.50 \\
$t+4$ &
  47.49 &
  18.25 &
  75.71 &
  77.32 &
  62.03 &
  26.49 &
  \textbf{78.02} &
  \textbf{79.91} &
  70.15 &
  57.41 &
  77.03 &
  76.47 \\ \bottomrule
\end{tabular}%
\end{table*}

\subsection{Incremental disease staging}

We apply our proposed approach to create an incremental disease staging agent. We design an incremental learning task for learning an Alzheimer's disease prediction model.

Alzheimer's disease staging is a non-trivial process with overlapping subjective categories. Due to the absence of a standard staging model for neurological diseases like AD, stage-wise labeled data may not be available at a single time point. We propose using task incremental learning to carry out source-free few-shot incremental updates to a base clinical model. to test our class incremental hypothesis, we aim to update a binary classification AD/HC model to predict intermediate stages of early mild cognitive impairment (EMCI) and late mild cognitive impairment (LMCI). To test our domain incremental hypothesis, we update the model using target data from a different domain (containing both shared and target private classes). 

We evaluate the method using Alzheimer's disease data from multiple domains, different populations, and different label sets. We use Alzheimer's disease-specific datasets in this experiment -- Alzheimer's Disease Neuroimaging Initiative
(Data used in the preparation of this article were obtained from the Alzheimer's Disease Neuroimaging Initiative (ADNI) database (adni.loni.usc.edu)) \cite{Cho2012} and Alzheimer's Disease Neuroimaging Initiative -- AIBL (Data was collected by the AIBL study group. AIBL study methodology has been reported previously \cite{ellis2009australian}).

We create a region of interest (ROI) image dataset using MRI images from ADNI and AIBL domains. The MRI images were pre-processed using a processing pipeline. Due to the relatively low number of samples in the MRI imaging dataset, we augment the dataset using the extracted ROIs from the input images \citep{liu2014multimodal, hosseini2016alzheimer}. For example, for the ADNI-1 dataset, we had 841 samples (200 healthy control data, 230 AD data, and 411 MCI data); after ROI augmentation, we had 3364 data instances.

We used ROI data from left and right Hippocampus regions and left and right temporal lobes. The extracted ROI patches had the dimension $(64\times64\times64)$. Individual ROI patches were labeled using the sample label from which they were extracted.

\begin{table*}[ht]
\centering
\caption{Incremental disease prediction task: comparison of our proposed method (TIDo) applied to Alzheimer's disease prediction with existing incremental learning, continual learning, and unsupervised domain adaptation methods for a few-shot labeled target (5\%), unlabelled target domain data and labeled source domain data. \textit{All(\%)} is the average accuracy for all the classes, \textit{Priv(\%)} is the average accuracy for private classes}
\label{adni_m3}
\begin{tabular}{@{}ccccccc@{}}
\toprule
      & \multicolumn{6}{c}{ADNI 1 (CN/AD) $\rightarrow$ ADNI 2 $\rightarrow$ AIBL $\rightarrow$ ADNI GO $\rightarrow$ ADNI 3} \\ \midrule
                          & \multicolumn{2}{c}{iCARL}       & \multicolumn{2}{c}{DANN-TPC}    & \multicolumn{2}{c}{CMA-TPC}              \\ \midrule
Index                     & All (\%)       & Priv (\%)      & All (\%)       & Priv (\%)      & All (\%)       & \textit{Priv (\%)}      \\ \midrule
$t+0$ & 90.01$\pm$3.08 & \textbf{88.01$\pm$1.29} & \textbf{93.44$\pm$2.25} & 54.91$\pm$1.04 & 80.52$\pm$1.70 & 52.28$\pm$1.27 \\
$t+1$                     & 84.91$\pm$2.58 & -              & \textbf{91.81$\pm$2.02} & -              & 83.78$\pm$2.01 & -                       \\
$t+2$                     & 80.92$\pm$3.66 & 71.22$\pm$3.02 & 71.25$\pm$4.89 & 57.05$\pm$2.28 & 82.67$\pm$1.18 & 67.82$\pm$4.60          \\
$t+3$                     & 78.32$\pm$4.81 & 70.10$\pm$4.12 & 73.72$\pm$4.21 & 56.81$\pm$2.56 & 84.19$\pm$1.29 & 63.91$\pm$5.67          \\ \midrule
\multicolumn{1}{l}{}      & \multicolumn{2}{c}{CIDA-C}      & \multicolumn{2}{c}{LwF-MC}      & \multicolumn{2}{c}{TIDo}                 \\ \midrule
\multicolumn{1}{l}{Index} & All (\%)       & Priv (\%)      & All (\%)       & Priv (\%)      & All (\%)       & \textit{Priv (\%)}      \\ \midrule
$t+0$                     & 89.32$\pm$1.17 & 82.90$\pm$3.16 & 72.91$\pm$0.98 & 56.68$\pm$1.10 & 91.42$\pm$0.79 & 86.91$\pm$1.22          \\
$t+1$                     & 90.76$\pm$2.01 & -              & 79.91$\pm$2.17 & -              & 90.08$\pm$2.48 & -                       \\
$t+2$                     & 89.62$\pm$1.57 & 90.22$\pm$2.78 & 77.67$\pm$2.11 & 57.52$\pm$4.17 & 90.69$\pm$2.20 & \textbf{92.82$\pm$1.71} \\
$t+3$                     & 90.32$\pm$2.89 & 88.10$\pm$2.11 & 78.10$\pm$1.07 & 59.24$\pm$1.70 & 92.12$\pm$0.91 & \textbf{91.14$\pm$0.88} \\ \bottomrule
\end{tabular}%
\end{table*}

\subsection{Discussion}

We proposed a source-free task incremental learning method for an agent to learn a task incrementally. We observed that our approach enabled an autonomous agent to learn a near-optimal target hypothesis with very low catastrophic forgetting for both class incremental and domain incremental applications. Since our approach is source-free, we have a very low memory complexity and can update a model incrementally using few-shot learning.

Our results show comparable or improved performance of our approach compared to class incremental learning (CIDA-C \citep{kundu2020class}). We show that our approach can achieve similar performance without storing past target training data. This reduces the memory complexity of our approach drastically.

We performed a comparative analysis of the task incremental problem using unsupervised domain adaptation, continual learning, and class incremental methods. \cite{rebuffi2017icarl} propose a supervised incremental learning approach that uses representation learning and learned class-wise exemplars from the input data. The authors updated the exemplars incrementally to learn using new classes and instances. Storage of class-wise exemplar instances and the need for labeled samples from both source and target domains for model upgrades make the approach unsuitable for scalable incremental learning. We eliminate the need to store exemplar instances by generating a distribution estimation and storing class-wise guides, thereby rendering our approach source-free. 

We compared our approach to continual manifold adaptation (CMA). CMA does not apply to open-set transfer learning settings. Hence, we learn a target private classifier (TPC) to achieve the task incremental task. Due to the few-shot configuration for target private instances, TPC risk is large. Table \ref{office31_tido} and \ref{adni_m3} show a high loss for target private classes, except $(t+2)^{th}$ iteration for Alzheimer's disease prediction (57.82$\pm$4.60\%) which is because the target private class (EMCI) is a sub-category of MCI, which has been observed by the classifier in the previous iterations ($t+0$, $t+1$) for related domains (ADNI 3 and ADNI 2).

\subsection{Ablation studies}

We will now explore the effect of different components of our proposed approach.

\begin{figure*}[ht]
    \centering
    \begin{subfigure}[t]{0.4\textwidth}
  	 	\begin{center}
		\includegraphics[width=\textwidth]{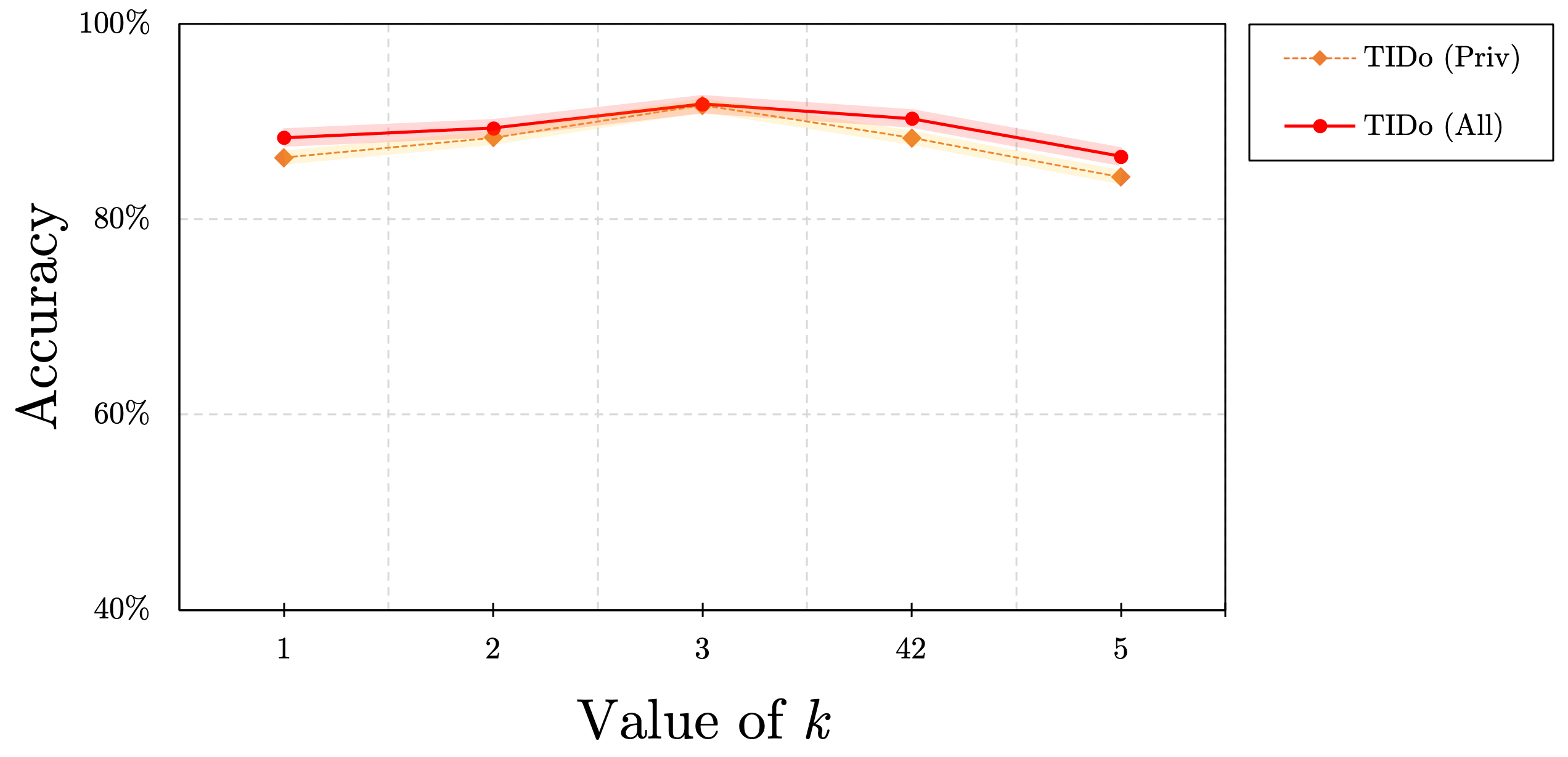}
		\caption{}
		\label{til_sigma}
		\end{center}
    \end{subfigure}%
    \begin{subfigure}[t]{0.4\textwidth}
  	 	\begin{center}
		\includegraphics[width=\textwidth]{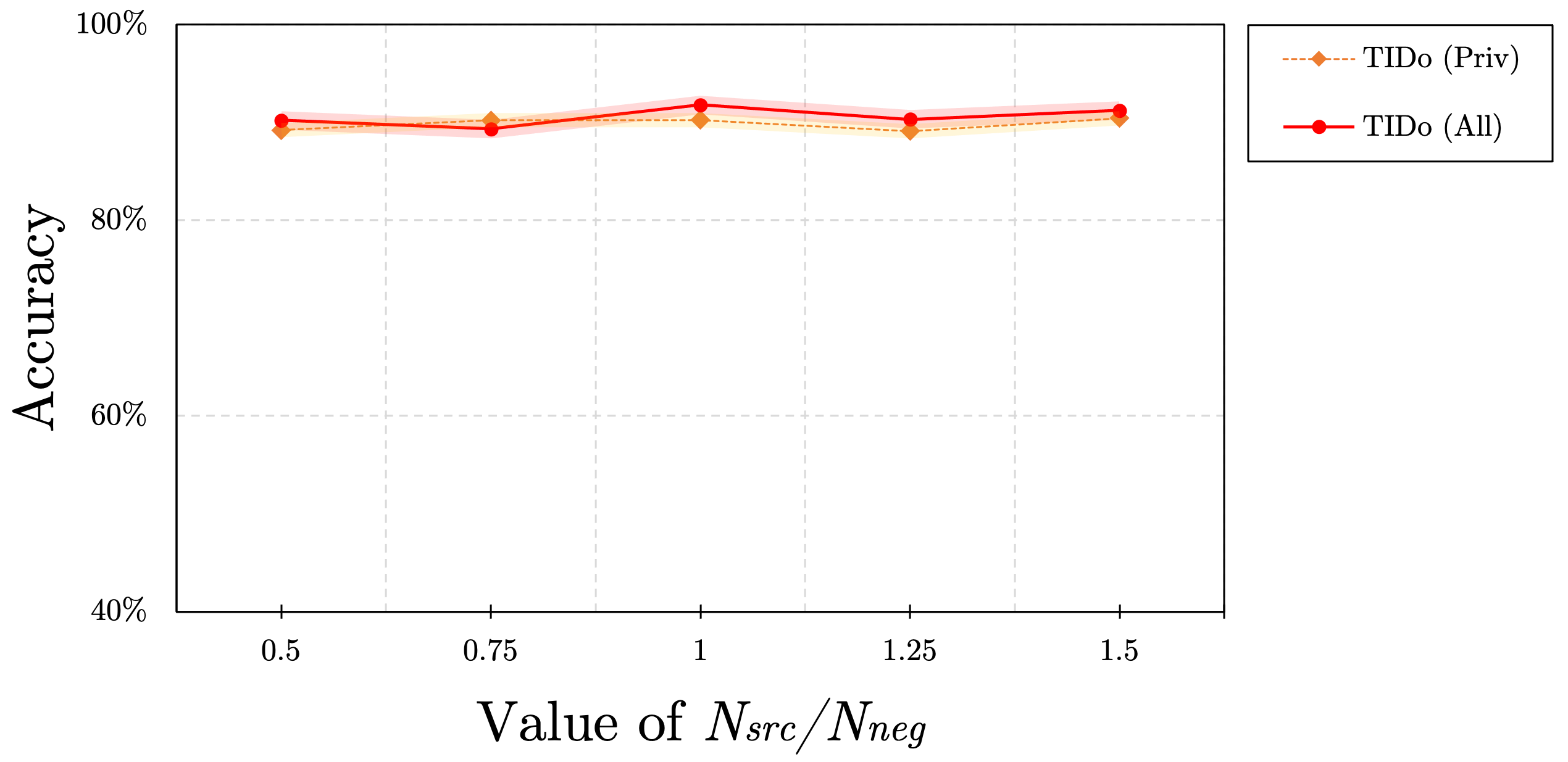}
		\caption{}
		\label{til_N}
		\end{center}
    \end{subfigure}%
    \vskip\baselineskip
    \begin{subfigure}[t]{0.4\textwidth}
  	 	\begin{center}
		\includegraphics[width=\textwidth]{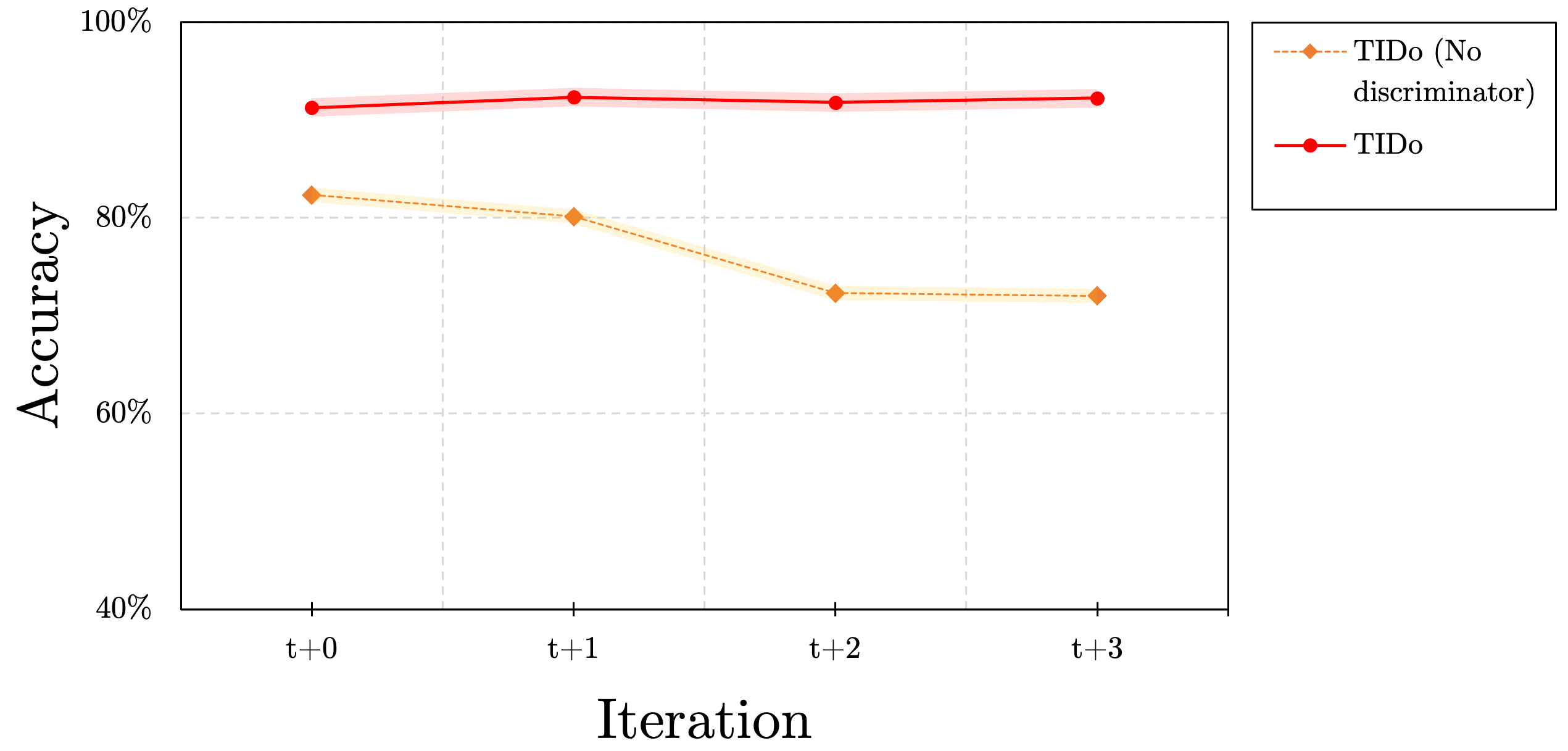}
		\caption{}
		\label{til_d}
		\end{center}
    \end{subfigure}%
    \begin{subfigure}[t]{0.4\textwidth}
  	 	\begin{center}
		\includegraphics[width=\textwidth]{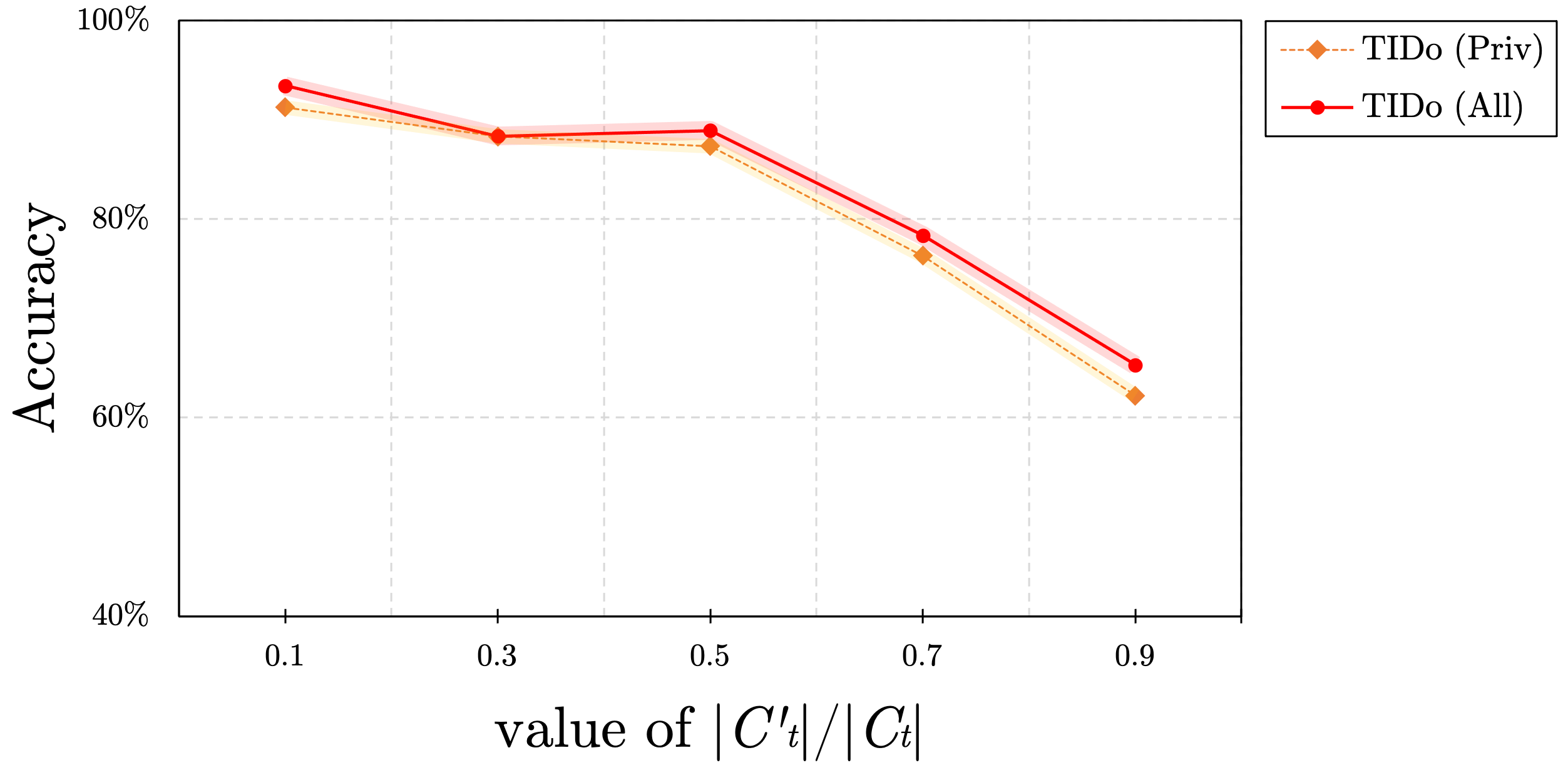}
		\caption{}
		\label{til_C}
		\end{center}
    \end{subfigure}%
    \caption{\textit{Sensitivity study results for task incremental  learning on incremental disease staging task (for AD):} (a) Effectiveness of Gaussian estimation and OOD sample prediction (Avg. accuracy\% for private and all target classes) (b) Data imbalance robustness for target task prediction (Avg. accuracy\%) (c) Effect of removal of discriminator in the foresighted model (d) Sensitivity on the ratio of private class sample size vs. all class sample size}
    \Description{\textit{Sensitivity study results for task incremental  learning on incremental disease staging task (for AD):} (a) Effectiveness of Gaussian estimation and OOD sample prediction (Avg. accuracy\% for private and all target classes) (b) Data imbalance robustness for target task prediction (Avg. accuracy\%) (c) Effect of removal of discriminator in the foresighted model (d) Sensitivity on the ratio of private class sample size vs. all class sample size}
\end{figure*}

\textbf{Effectiveness of Gaussian estimation and OOD sample prediction:} Similar to previous approaches, we analyze the sensitivity of the hyper-parameter $k$ to observe the effects of modifying the labeling criteria for negative samples in the foresighted learning stage. To verify that our Gaussian estimates are accurate, we empirically tested the efficiency of the assumed confidence interval (3-$\sigma$). Figure \ref{til_sigma} shows that 3-$\sigma$ provided the maximum predictive accuracy and best captured the source distribution characteristics.
    
\textbf{Effect of balancing source and target unlabelled data:} We used a balanced source ($N_{src}$) and target ($N_{neg}$) domain dataset to train our baseline model. We test the robustness of our model to imbalanced data by varying the $N_{src}/N_{neg}$ ratio by $\pm0.5$. We measure the sensitivity of the source and target domain ratio in figure \ref{til_N} and observe that the proposed approach is robust against data imbalance. 
    
\textbf{Challenging one-shot learning:} We observe the efficiency of our incremental learning approach by varying the ratio of samples in the target private classes to the number of shared class samples ($|C_{t}^{'}|/|C_{t}$). Figure \ref{til_C} shows the sensitivity of this ratio. Even though a larger number of target private samples improves the accuracy of private guides and private class prediction, prediction accuracy reduces due to the inability of the target classifier to converge under less target shared class data. 
    
\textbf{Effect of class separation loss:} We carry out the ablation study by removing the class separation loss. We learn the post-increment accuracy of the target domain classifier without applying the class separation loss ($\mathcal{L}_{s1}$). We observe that the average prediction accuracy without the loss minimization was 83.45\% compared to 92.12\% using the class separation loss.

\section{CONCLUSION}

In this work, we proposed an approach for an autonomous agent to learn a task incremental learning model in a non-stationary environment. We explored a one-shot learning approach to reduce the need for collecting labeled data to incrementally update a model. Using a source-free approach, we were able to learn aligned target private prototype guides and learn with very few target-labeled samples. One of the limitations of our approach is the possibility of overfitting after a given number of incremental iterations. We aim to address this limitation in our future work by exploring selective forgetting using recurrent network-based approaches. Another possible limitation of this work would be the use of Gaussian estimates to generate replay memory representative samples. We aim to explore adversarial methods to generate representative samples in our future work.

\bibliographystyle{ACM-Reference-Format}
\bibliography{main}

\end{document}